%% file: CameraReady2027.tex
\documentclass[letterpaper]{article} 
\usepackage{aaai2027}  
\usepackage[hyphens]{url}  
\usepackage{graphicx} 
\usepackage{natbib}  
\usepackage{caption} 
\usepackage{amsmath}
\usepackage{cleveref} 
\usepackage{algorithm}
\usepackage{algpseudocode}
\usepackage{makecell}
\usepackage{newfloat}
\usepackage{listings}
\DeclareCaptionStyle{ruled}{labelfont=normalfont,labelsep=colon,strut=off} 
\floatstyle{ruled}
\newfloat{listing}{tb}{lst}{}
\floatname{listing}{Listing}

\usepackage{booktabs}

\usepackage{amssymb}
\usepackage{mathrsfs}
\usepackage{multirow} 

\title{Driver2Map: Imitating Human Driving \\for Online High-Definition Map Construction}
\author{
    Pan Yin\equalcontrib, Runtian Xia\equalcontrib,
    Weisong Kuang,
    Kaiyu Li\corresponding, Cong Zhao, Xiangyong Cao\corresponding
}
\affiliations{

    Xi'an Jiaotong University\\
}

\nocopyright

\begin{document}

\maketitle


\begin{abstract}
High-definition (HD) maps are essential for autonomous driving systems. In constructing such maps,  onboard multi-view camera images, standard-definition maps and satellite images provide crucial information. However, due to the modality and perspective differences among these data sources, existing methods often struggle to effectively align and fuse them, making online HD map construction still challenging. To address these issues, we propose \textbf{Driver2Map}, an online HD map construction model inspired by human drivers. Unlike existing HD map construction models that utilize only two modalities, our Driver2Map can 
simultaneously exploit three modalities. Specifically, we propose a ``two-stage alignment'' strategy to reduce spatial misalignment across different modalities. Additionally, we introduce ``Pose-Guided BEV Fusion'', a BEV (bird's-eye-view) generation module that leverages camera pose information to adaptively weight multi-view features, thereby effectively suppressing cross-view feature overlap during BEV generation. Also, we design a ``Pretrained Prior for Map Refinement'' module to refine the initial prediction by learning map structure priors, thus improving the HD map prediction under dynamic occlusions. Extensive experiments demonstrate that Driver2Map outperforms existing methods on both IoU and AP metrics. Our code is available at \url{https://github.com/UserBits/Driver2Map}.


\end{abstract}


\section{Introduction}

\label{sec:intro}
\begin{figure}[t]
    \centering
    \includegraphics[height=4.1cm]{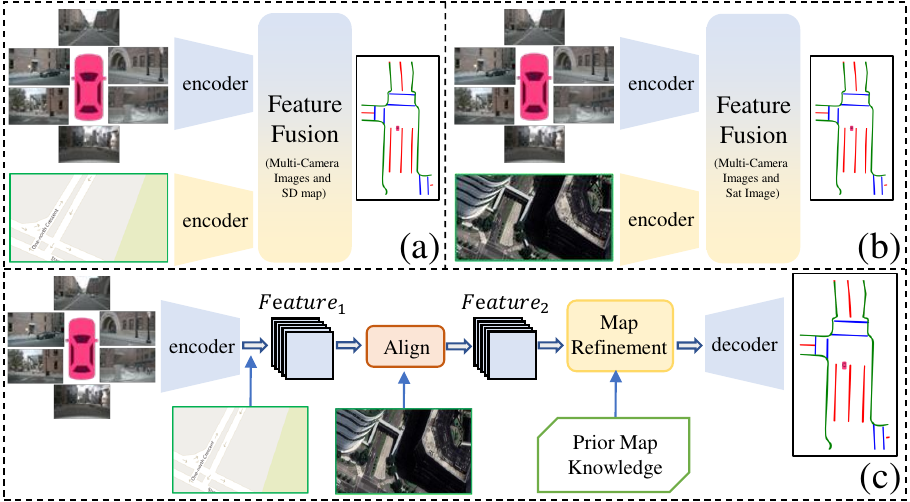}
    \caption{Comparison between our Driver2Map and other models.  (a) and (b) show the HD map construction process using SD map and satellite image as prior respectively, and (c) illustrates our proposed method Driver2Map. }
    \label{fig:introduction}
\end{figure}

High-definition (HD) maps provide detailed geometric and semantic information, including lane dividers, road boundaries and pedestrian crossings~\cite{caesar2020nuscenes}, serving as essential priors for autonomous driving tasks such as localization and path planning~\cite{wen2020urbanloco,charroud2024localization}. However, dynamic road environments require HD maps to be continuously updated~\cite{zhou2019spatial}. Traditional offline map construction pipelines~\cite{orbslam2,liosam, arora2023static} rely on costly and time-consuming manual annotations. Therefore, recent studies construct HD maps online from cost-effective vehicle-view images~\cite{liu2023vectormapnet, liao2023maptr}. However, the perspective discrepancy between vehicle-view images and the final HD maps often causes positioning errors and geometric distortions.
To alleviate these issues, recent studies incorporate additional priors such as standard-definition (SD) maps and satellite images to complement onboard observations (see Figure~\ref{fig:introduction}). HRMapNet~\cite{zhang2024hrmapnet} and Topo-SD~\cite{yang2024toposd} use SD map to incorporate road topology as prior, but still lack road details. Meanwhile, SatforHDMap \cite{gao2024satforhdmap} integrates satellite images to compensate for vehicle-based observation blind spots and recover map details. However, lacking the road topological information, such methods often struggle to recover accurate road geometries in regions occluded by buildings and other objects in the satellite image. Nevertheless, jointly incorporating SD maps' and satellite images' prior information into the HD map construction process remains challenging. On the one hand, existing prior information for SD map and satellite image are typically defined in different coordinate systems, leading to the accumulation of errors in the alignment between different data sources~\cite{li2024geolocalization}. On the other hand, onboard multi-view images, SD maps, and satellite images differ significantly in both modality and perspective~\cite{yang2024toposd}, making effective cross-modal information fusion inherently difficult.


\begin{figure*}[t]
  \centering
  \includegraphics[height=6.2cm]{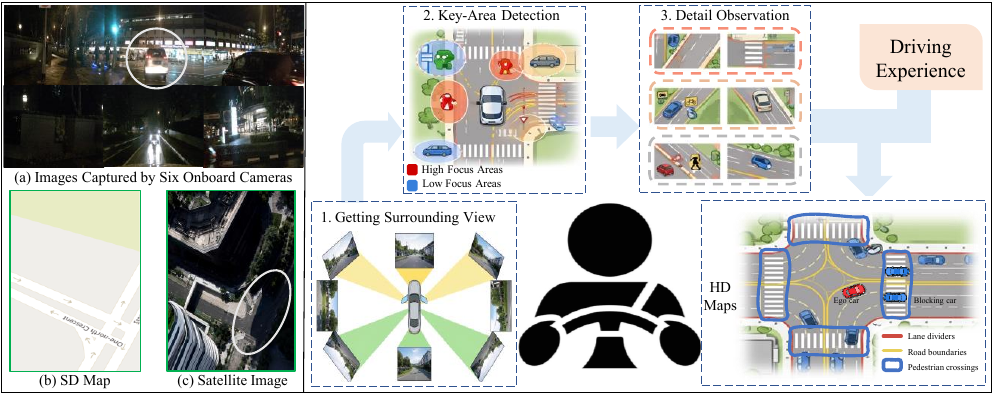}
  \caption{Motivation of Driver2Map: multi data source limitations (left) and human drivers' map-building process (right). The left hows the limitations of different data sources. Onboard camera images are unreliable under poor illumination; SD maps provide only coarse road topology; both satellite image and onboard camera images can be occluded. The right illustrates human drivers' map-building process that inspires Driver2Map: first perceiving the surroundings, then identifying key road areas and performing more detailed observations, and finally using driving experience to infer road elements occluded by dynamic objects.}
  
  \label{fig:driver}
\end{figure*}


To address the above issues, we propose \textbf{Driver2Map}, an online HD map construction framework that jointly exploits onboard multi-view camera images, local SD map and satellite image. 
First, to resolve the inconsistency across coordinate systems in different priors, we propose a two-stage alignment strategy. Based on the vehicle's location and orientation, the strategy employs affine transformations to account for scaling, translation, and rotation discrepancies, thereby aligning the priors within a unified coordinate system and jointly constructing the matching SD and satellite map tiles. 

Then, to better leverage the two types of prior information, we progressively incorporate them by simulating the map-building process of human drivers. As illustrated in Figure~\ref{fig:driver}, human drivers typically perceive their surroundings to form an initial understanding of the road environment, then focus on key areas to perform structural recognition, and finally improve the information using historical driving experience when encountering occlusions. Inspired by this, Driver2Map first aggregates features from onboard multi-view images to construct an initial scene representation. Since the vehicle-view features are not aligned with the target BEV representation, we further transform them into the BEV space. Existing methods \cite{huang2021bevdet,philion2020lift} often fuse features from different cameras indiscriminately during BEV generation, allowing irrelevant views to interfere with the target BEV region (\textit{e.g.}, features from the rear view may affect the front BEV region). To address this, we propose a ``Pose-Guided BEV Fusion'' module, which uses vehicle pose information to guide cross-view feature alignment and aggregation. Subsequently, Driver2Map incorporates the SD map and satellite image as priors. The SD map provides structured topological cues to identify key road structure, while the satellite image further enriches fine-grained geometric details, enabling more accurate and complete HD maps.

Finally, to address incomplete map construction caused by dynamic obstructions such as vehicles and pedestrians, we draw inspiration from the inference strategy of human drivers and design a module, ``Pretrained Prior for Map Refinement''. 
By learning structural priors from HD maps through pretraining, the module effectively refines and completes the initial predictions. Experiments on the nuScenes dataset demonstrate that under different BEV ranges and across multiple semantic categories, our Driver2Map outperforms existing methods in terms of IoU and AP metrics.

In summary, our main contributions are three-fold: 
\begin{itemize}
    \item We propose Driver2Map, a novel online HD map construction framework that jointly leverages vehicle-view images, SD maps, and satellite images, with a two-stage alignment strategy for cross-modal correspondence. 
    \item We introduce a ``Pose-Guided BEV Fusion'' module that leverages relative camera poses to guide cross-view feature aggregation and BEV generation.
    \item We propose a ``Pretrained Prior for Map Refinement'' module to refine incomplete predictions caused by dynamic occlusions.
\end{itemize}


\section{Related Works}

\subsection{Online HD Map Construction}


Recent advances in online HD map construction have significantly reduced the cost and latency of traditional offline map generation. Early methods, such as Lift-Splat-Shoot~\cite{philion2020lift} and VectorMapNet~\cite{liu2023vectormapnet}, construct HD maps solely from onboard multi-view camera images, enabling lightweight deployment but suffering from performance degradation under occlusions and poor lighting conditions. Therefore, methods like HDMapNet~\cite{li2022hdmapnet}, BEVFusion~\cite{liang2022bevfusion} and SuperFusion~\cite{dong2024superfusion} incorporate LiDAR point clouds to provide explicit spatial distance information, at the expense of increased sensor cost. Subsequent studies further exploit temporal information to improve map consistency. For example, BEVFormer~\cite{huang2023bevformer}, StreamMapNet~\cite{yuan2024streammapnet}, and HRMapNet~\cite{zhang2024hrmapnet} aggregate historical BEV features to enhance spatial-temporal perception, although their performance relies on effective long-term feature propagation. 
More recently, researchers have introduced external priors to compensate for the limitations of onboard perception. Topo-SD~\cite{yang2024toposd},  SatforHDMap~\cite{gao2024satforhdmap} and SDTagNet~\cite{immel2026sdtagnet} exploit SD map or satellite image to provide road structural and geometric priors. However, they still struggle with cross-modal alignment and fusion. 


\subsection{From Multi-View Images to BEV}


Existing methods for transforming multi-view images into BEV representation can be divided into two categories: geometry-based methods and attention-based methods. Geometry-based methods, such as Lift-Splat~\cite{philion2020lift} and BEVDet~\cite{huang2021bevdet}, estimate depth distributions to project image features into 3D space before voxelization. Though effective, they rely on depth estimation and are computationally expensive. In contrast, attention-based methods, such as BEVFormer~\cite{huang2023bevformer}, directly model cross-view spatial correlations without explicit depth estimation. Subsequent works, including Fast-BEV~\cite{huang2023fast}, SparseBEV~\cite{liu2023sparsebev}, and BEVStereo~\cite{li2023bevstereo}, further improve efficiency, depth estimation accuracy and robustness. Despite these advances, existing methods generally ignore relative pose relationships among different camera views during feature fusion, leading to cross-view interference and loss of fine-grained details.

\subsection{SD Map and Sat Image for Map Construction}

SD maps and satellite images provide complementary priors for map construction~\cite{jin2005integrated}. SD maps typically encode coarse, structural information~\cite{jin2005integrated}, while satellite images offer a global bird's-eye view and fine-grained geometric details~\cite{burke2021using, abburu2015satellite}. Recently, SamRoad~\cite{hetang2024segment} employs SAM~\cite{kirillov2023segment} to extract road centerlines, while SamRoad++~\cite{yin2025towards} introduces an extended-line strategy to alleviate occlusions in satellite views. Furthermore, \cite{huang2025satmaptr, mazumder2026satmap} also attempt to enhance the detail quality of HD maps with satellite images.
In parallel, SamMaps~\cite{van2025sam} and Loop-MapNet~\cite{tang2025loop} incorporate SD map priors to improve road extraction accuracy.

Motivated by this, we incorporate both SD maps and satellite images as external priors for HD map construction. Specifically, SD maps provide coarse topological guidance for identifying key road structures, while satellite images offer lane-level fine cues to recover map details.

\section{Method}

{
\begin{figure*}[!t]
  \centering
  \includegraphics[height=8cm]{
  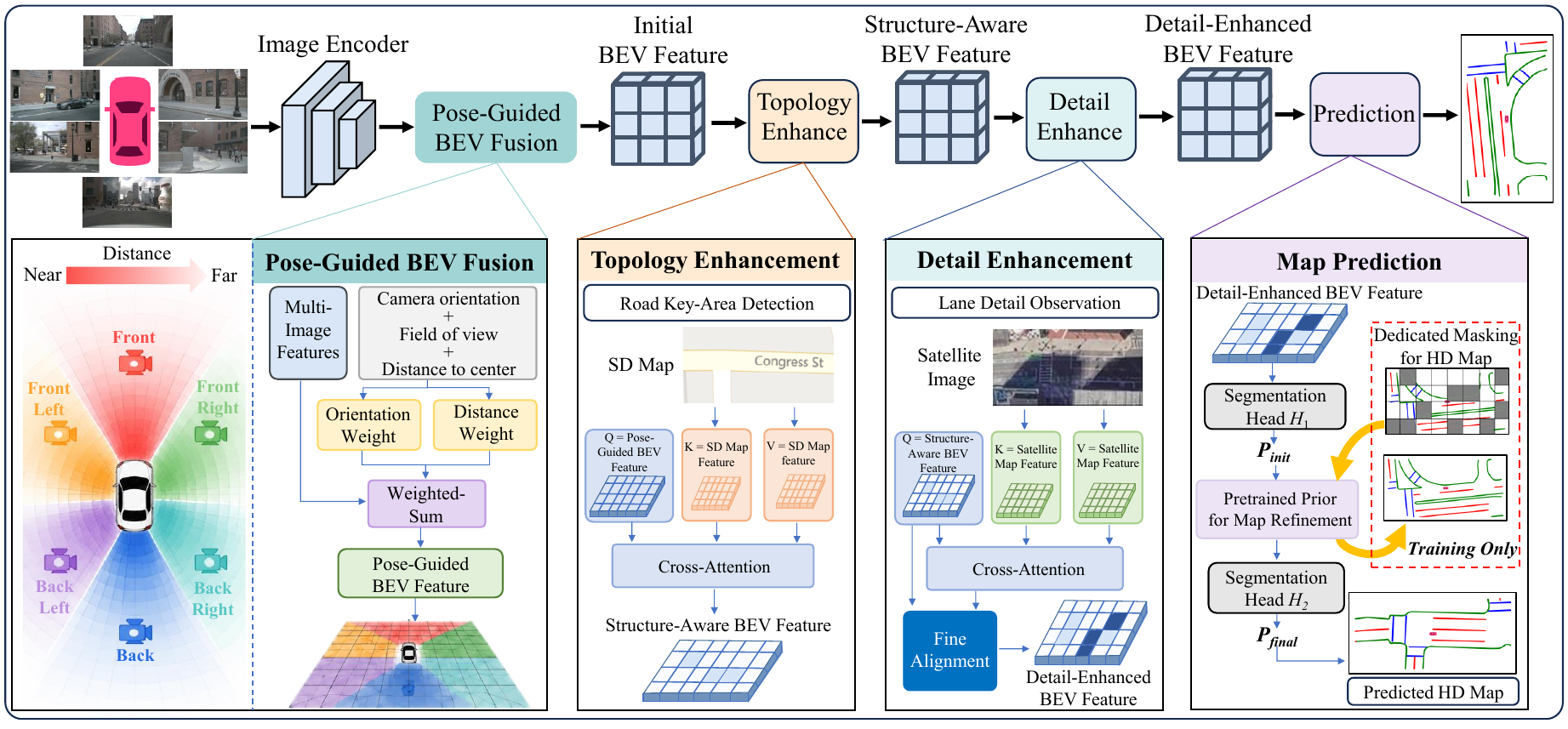}
  \caption{Overview of Driver2Map. Inspired by the process of map construction by human drivers, our model follows four stages: surrounding perception with onboard multi-view images; key road area focusing with SD maps; detail enhancement with satellite images; and restoring missing structures under dynamic occlusion with ``Pretrained Prior for Map Refinement'' module.}
  \label{fig:method}
\end{figure*}
}

\subsection{Overview}

As shown in Figure~\ref{fig:method}, inspired by the map construction process of human drivers, Driver2Map progressively aggregates onboard multi-view images, SD map, and satellite image. First, 
multi-view features are extracted by an EfficientNet-B0~\cite{tan2019efficientnet} encoder 
and transformed into the BEV space through a proposed ``Pose-Guided BEV Fusion'' module (Section~\ref{sec:image-to-bev}), generating the initial BEV feature.

As shown in Figure~\ref{fig:driver}, human drivers tend to focus more on road areas during driving. 
Motivated by this, we design a topology enhancement module that first uses a ResNet~\cite{he2016resnet} encoder to extract features from SD maps (which provide road topology), and then fuses the extracted features with the initial BEV feature via an attention mechanism to obtain the structure-aware BEV feature. Since SD maps lack fine-grained lane details, a detail enhancement module further integrates the satellite features extracted by another ResNet encoder, producing the detail-enhanced BEV feature. Finally, a CNN segmentation head is used to predict the initial HD map $P_{\text{init}} \in \mathbb{R}^{ (\text{num\_classes}+1) \times H \times W }$, where $\text{num\_classes}=3$ denotes the number of semantic categories excluding the background class.



Finally, to recover map elements missing due to dynamic occlusions, we propose the ``Pretrained Prior for Map Refinement'' module (Section~\ref{sec:fine-tune}), which learns HD map structure priors in advance to reasonably refine the $P_{init}$ and to output the final HD map predictions $P_{final}$.



\subsection{Pose-Guided BEV Fusion}
\label{sec:image-to-bev}

To align with the bird’s-eye view (BEV) of the final HD map, we first transform onboard multi-view images into a BEV feature representation. However, previous methods typically ignore the differences in orientation and distance across onboard cameras, which often leads to misleading information during the fusion process~\cite{li2023bevdepth, huang2023bevformer}. For example, image information from the front of the vehicle can mislead the BEV feature representation of the back. Therefore, we propose ``Pose-Guided BEV Fusion'', which addresses the above issues by leveraging the pose differences across different onboard cameras.



As shown on the left of Figure~\ref{fig:method}, the front region in the BEV map should mainly rely on the front camera, supplemented by the front-left and front-right cameras. Therefore, our ``Pose-Guided BEV Fusion'' leverages a weighted-sum scheme that assigns each BEV region to its dominant camera view, while enabling regions near camera view boundaries to incorporate information from adjacent cameras with smooth transitions. First, 
based on each camera's orientation angle $\theta_{\text{cam}}^i$ and horizontal field of view $\theta_{\text{view}}^i$, we partition the BEV space into six sectors labeled ${i=0,1,\dots,5}$, with each sector corresponding to an independent camera view. 
For camera $i$, its weight is determined by both distance and orientation.



\begin{itemize}
    \item \textbf{Distance Weight.} For sector $i$ in the BEV, 
    let $d^i(x,y)$ denote the Euclidean distance between point $(x,y)$ and the geometric center of sector $i$. The distance weight contributed by camera $i$ at location $(x,y)$ is defined as:
    
    \begin{equation}
    W_{\text{dis}}^i(x,y) =
    \frac{1}{d^i(x,y)}, 
    \label{eq:w_dis}   
    \end{equation}

    \item \textbf{Orientation Weight.} Each camera $i$ (or each sector $i$) has a central viewing direction $\theta_{\text{cam}}^i$. For any point $(x,y)$ inside BEV, let $\theta^i(x,y)$ denote the direction from the BEV center to $(x,y)$. The orientation weight at location $(x,y)$ contributed by camera $i$ is defined as:

\begin{equation}
\begin{aligned}
W_{\mathrm{ori}}^i(x,y)=
\begin{cases}
\cos(\delta_i(x,y)), & \delta_i(x,y)\le \tfrac{1}{2}\theta_{\mathrm{view}}^i,\\
0, & \mathrm{otherwise}.
\end{cases}
\end{aligned}
\label{eq:w_ori}
\end{equation}

    where $\delta_i(x,y)=|\theta^i(x,y)-\theta_{\text{cam}}^i|$. Equation~\ref{eq:w_ori} assigns lower weights to locations with larger angular deviations and suppresses regions outside the camera's field of view.


\end{itemize}
The final weight at location $(x,y)$ is obtained by normalizing the weighted sum across all six cameras:

\begin{equation}
\begin{aligned}
W_{\text{cam}}^i(x,y)
&=
\frac{
\alpha W_{\text{dis}}^i(x,y)+\beta W_{\text{ori}}^i(x,y)
}{
\sum_{j=0}^{5}
\left(\alpha W_{\text{dis}}^j(x,y)+\beta W_{\text{ori}}^j(x,y)\right)
}
\end{aligned}
\label{eq:w_cam}
\end{equation}
where $\alpha$ and $\beta$ are the coefficients balancing the distance and orientation weights, respectively.

Finally, the 6-view camera image features \( F_{\text{cam}} \in \mathbb{R}^{6 \times H \times W \times C} \) are then aggregated using the calculated per-camera weights \( W_{\text{cam}}^i(x,y) \). To this end, we first isolate the \( i \)-th channel group corresponding to the \( i \)-th camera perspective within \( F_{\text{cam}} \), denoted as \( F_{\text{cam}}^i \in \mathbb{R}^{H \times W \times C} \) (where \( i \in \{0,1,\dots,5\} \) indexes the six cameras). The original BEV feature \( F_{\text{bev}} \in \mathbb{R}^{H \times W \times C} \) is computed as the weighted sum of these channel-grouped features:
{
\setlength{\intextsep}{0.5em} 
\setlength{\abovedisplayskip}{0.5em} 
\setlength{\belowdisplayskip}{0.5em} %
\begin{equation}
F_{\text{bev}}(x,y,c) = \sum_{i=0}^{5} W_{\text{cam}}^i(x,y) \cdot F_{\text{cam}}^i(x,y,c)
\end{equation}
}


\subsection{Two-Stage Alignment}


\label{sec:prior-map}

\label{sec:data-pre}


As shown in Figure~\ref{fig:keypoint}, spatial offsets across maps caused by translation, rotation, and scale differences make it challenging to obtain matched SD map and satellite image priors for each driving sample. To address this, we propose a two-stage alignment strategy consisting of coarse alignment during data preparation and fine alignment within the model.


\subsubsection{Coarse Alignment.} \label{sec:coar}
\begin{figure}[t]
    \centering
    \includegraphics[height=4cm]{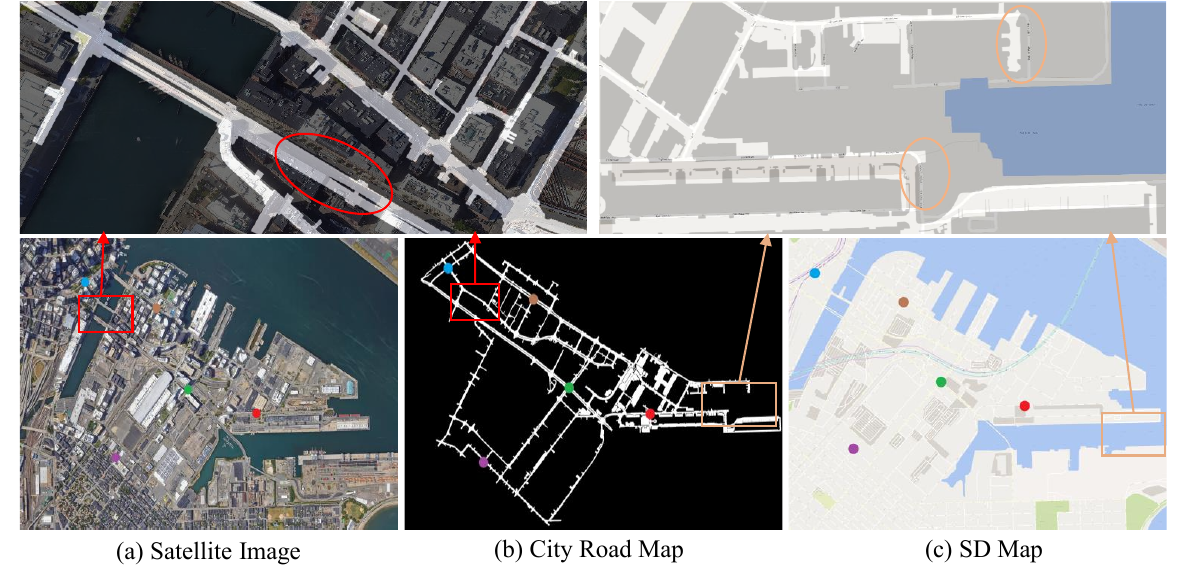}
    \caption{Visualization of offsets across different maps and keypoint selection. The top shows road misalignment across maps caused by translation and rotation; the bottom illustrates the keypoint selection used for coarse alignment.}
    \label{fig:keypoint}
\end{figure}

Geometric misalignment across different maps may lead to mismatched priors for each driving sample (see Figure~\ref{fig:keypoint}). To alleviate this, we construct SD map and satellite image priors aligned with the nuScenes road map via a general coarse alignment method.


\input{CoarseAlignment}

An example of the construction procedure of SD map and satellite map tiles of Boston Seaport for nuScenes is shown in Algorithm~\ref{alg:coarse_alignment}. First, we obtain the large-scale and high-resolution SD map and satellite image of this area from Google Maps. 
Then, to simultaneously process the scaling, translation, and rotation across the nuScenes road map, the SD map and the satellite image, we apply an affine transformation to convert the coordinate system of the SD map or satellite image to that of the nuScenes road map:
\[
\mathrm{FitAffine}(\mathcal{K}) \Rightarrow
\left\{
\begin{aligned}
x' &= ax + by + e \\
y' &= cx + dy + f
\label{eq:aff}
\end{aligned}
\right.
\]

\noindent
where $(x, y)$ denotes the coordinate of a point in the SD map or satellite image, and $(x', y')$ represents the corresponding point in the road map. We manually select $N$ pairs of keypoints $(x_i, y_i)$ and $(x'_i, y'_i)$ ($i = 1, 2, \dots, n$) and use the least squares method to accurately fit the parameters $a \sim f$ in Equation~\ref{eq:aff}. It is notable that the mapping from the road map to the SD map and that to the satellite image are estimated independently, resulting in different affine transformation parameters. Using the transformed coordinates as the center, we perform a rotation of the large-scale SD map (or satellite image) according to the vehicle's current heading, and crop $100 \times 200$ pixel map tiles.




\subsubsection{Fine Alignment.} Coarse alignment mitigates most of the translation, rotation, and scale errors. However, 
slight residual misalignments still remain. Therefore, we further introduce a learnable fine alignment module in the model stage.


Within the Detail Enhancement module (Figure~\ref{fig:method}), we further perform fine alignment to reduce residual misalignment before feature fusion. Specifically, the cross-attention feature $F_{\text{ca}}$ and the structure-aware BEV feature $F_{\text{bev}}$ are first concatenated and passed through convolutional layers to predict a dense 2D flow-field $\Delta \in \mathbb{R}^{H \times W \times 2}$, where each vector denotes a pixel-wise coordinate offset. The predicted flow field is then used to warp $F_{\text{ca}}$ via differentiable bilinear sampling:

\begin{equation}
\widetilde{\mathbf{F}}_{ca}(p)=\mathbf{F}_{ca}(p+\Delta(p))
\end{equation}
where bilinear interpolation is adopted for sampling at non-integer coordinates. The final detail-enhanced BEV feature $\widetilde{\mathbf{F}}_{ca}$ is subsequently used for the following map prediction.



\subsection{Pretrained Prior for Map Refinement}
\label{sec:fine-tune}

To mitigate the issue of dynamic occlusions during driving, we design a ``Pretrained Prior for Map Refinement'' module inspired by human experience-based inference. This module consists of a masked auto-encoder (MAE)~\cite{he2022masked,dosovitskiy2020image} and a CNN segmentation head ${H}_2$, which is similar to the earlier head ${H}_1$.

During pretraining, we randomly mask HD maps from the nuScenes training set to simulate information loss caused by dynamic objects during driving. Since HD map construction typically involves three semantic categories, i.e., \textit{divider}, \textit{crossing}, and \textit{boundary}, which exhibit highly imbalanced spatial distributions, we adopt a category-wise patch masking strategy. Furthermore, if a patch is fully masked across all three semantic classes, we replace it with random partial-pixel masking to preserve partial supervision, enabling the module to acquire sufficient reconstruction capability rather than
losing all training signals. This pretraining can
 benefit the subsequent fine-tuning stage for HD map refinement.



To achieve a more balanced category learning, we dynamically adjust class weights during pretraining. Let $\mathrm{IoU} \in \mathbb{R}^3$ be computed between the output of the refinement module and the corresponding ground truth (GT), evaluated separately for \textit{divider} ($i=0$), \textit{crossing} ($i=1$), and \textit{boundary} ($i=2$) after each epoch. The weight for class $i$ in next epoch's semantic loss is:
\begin{equation}\label{eq:w_loss}
\begin{split}
w_{\mathrm{loss}}^i
&= 3 \cdot \mathrm{norm}\!\left(\exp(1-\mathrm{IoU}_i)\right) \\
&\quad \cdot \exp\!\left(1-\mathrm{mean}(\mathrm{IoU})\right).
\end{split}
\end{equation}


\noindent where $3 \cdot \mathrm{norm}(\cdot)$ ensures normalization so that the sum of the three weights equals $3.0$, preserving the original scale. The factor $\exp(1-\mathrm{mean}(\mathrm{IoU}))$ acts as a penalty coefficient: when the mean IoU across all classes is low, the absolute loss values increase, accelerating learning. 
After pretraining, the HD map prior module is integrated into the backbone model for fine-tuning. During fine-tuning, the module takes $P_{\mathrm{init}}$ as input and outputs the refined HD map prediction $P_{\mathrm{final}}$.   


\section{Experiments}

\subsection{Settings}

We train and evaluate Driver2Map on the nuScenes~\cite{caesar2020nuscenes} dataset. The model is trained for three stages on 4$\times$RTX4090 GPU. We first train the main backbone without the ``Pretrained Prior for Map Refinement'' (PPMR) module, and then pretrain PPMR independently using the dynamic category-wise loss weights in Eq.~\ref{eq:w_loss}. Finally, PPMR is integrated with the backbone and fine-tuned while the backbone remains frozen. We use AdamW as the optimizer, with learning rates of $1\times10^{-3}$, $8\times10^{-4}$, and $5\times10^{-5}$ for the three stages, respectively. The balancing coefficients in Eq.~\ref{eq:w_cam} are set to $\alpha:\beta=1:2$. For Algorithm~\ref{alg:coarse_alignment}, we select $N=50$ pairs of keypoints for each region. Predictions are evaluated using Intersection over Union (IoU)~\cite{rezatofighi2019generalizediou,li2022hdmapnet} and Average Precision (AP)~\cite{li2022hdmapnet}.

\begin{table}[t]
\centering

\setlength{\tabcolsep}{2pt}
\begin{tabular}{c|c|c|c|c|c}
\hline
\multirow{2}{*}{Method} & \multirow{2}{*}{Input} & \multicolumn{3}{c|}{AP(\%)} & \multirow{2}{*}{mAP} \\
\cline{3-5}
                        &                           & Div. & Cross. & Bound. &                        \\
\hline

\hline
FastMap\shortcite{hu2025fastmap}	& C	& 54.5	& 44.6	& 41.6	& 46.9
\\
\hline
PivotNet\shortcite{ding2023pivotnet}	& C	 & 53.6	& 43.4	& 50.5	& 49.2
\\
\hline
MapTRv2\shortcite{liao2025maptrv2}                 & C                         & 55.1    & 46.9     & 54.9     & 52.3                   \\
\hline
DAMap\shortcite{dong2025damap} & C & 55.6 & 62.2 & 59.4 & 59.1 \\
\hline
\makecell[c]{VectorMapNet+\\ SatforHDMap\shortcite{gao2024satforhdmap}} & C+Sat                     & 51.4    & 42.5     & 44.1     & 46.0                   \\
\hline

\makecell[c]{MapTR+\\ SatforHDMap\shortcite{gao2024satforhdmap}}       & C+Sat                  & 55.3    & 47.2     & 55.3     & 52.6                   \\
\hline
P-MapNet\shortcite{jiang2024pmapnet}                & C+SD                      & 50.9    & 43.7     & 53.5     & 49.4                   \\
\hline

SDTagNet\shortcite{immel2026sdtagnet}  & C+SD & 56.4  & 57.7 & 60.8  & 58.3  \\
\hline

Driver2Map(ours)        & C+Sat                     & 53.8    & 61.5     & 65.4     & 60.2                   \\
\hline
Driver2Map(ours)   & C+SD & 52.7 & 60.6 & 65.7 & 59.7 \\
\hline
Driver2Map(ours)        & C+SD+Sat                  & \textbf{57.1}    & \textbf{64.0}     & \textbf{67.5}     & \textbf{62.9}                   \\
\hline
\end{tabular}
\caption{Results on the nuScenes validation set under a $60\,\text{m} \times30\,\text{m}$ BEV range in AP metrics. \textbf{C}, \textbf{SD}, and \textbf{Sat} denote onboard multi-view camera images, standard-definition maps, and satellite images, respectively. }
\label{tab:ap}
\end{table}

\begin{table}[h]
\centering

\small
\setlength{\tabcolsep}{3pt}
\renewcommand{\arraystretch}{1.08}

\begin{tabular}{c|c|c|c|c|c}
\hline
\multirow{2}{*}{Method} & \multirow{2}{*}{Input} & \multicolumn{3}{c|}{IoU(\%)} & \multirow{2}{*}{mIoU} \\
\cline{3-5}
 &  & Div. & Cross. & Bound. &  \\
\hline

\multicolumn{6}{c}{\textbf{\rule{0pt}{1.1em}BEV range: $60\,\text{m}\times30\,\text{m}$}} \\
\hline

\makecell[c]{HDMapNet+\\ SatforHDMap\shortcite{gao2024satforhdmap}} & C+Sat & 47.2 & 40.1 & 46.9 & 44.7 \\
P-MapNet\shortcite{jiang2024pmapnet} & C+SD & 44.3 & 23.3 & 43.8 & 37.1 \\
SuperFusion\shortcite{dong2024superfusion} & C+L & 47.9 & 37.4 & 58.4 & 47.9 \\
CRFusion\shortcite{guan2026crfusion} &
C+L & 48.2 & 36.5 & \textbf{59.8} & 48.1 \\
Driver2Map(ours) & C+Sat & 51.5 & 48.0 & 50.4 & 50.0 \\
Driver2Map(ours) & C+SD+Sat & \textbf{56.7} & \textbf{53.4} & 51.9 & \textbf{54.0} \\
\hline

\multicolumn{6}{c}{\textbf{\rule{0pt}{1.1em}BEV range: $120\,\text{m}\times60\,\text{m}$}} \\
\hline
\makecell[c]{HDMapNet+\\ SatforHDMap\shortcite{gao2024satforhdmap}} & C+Sat & 41.2 & 30.1 & 42.9 & 38.1 \\
P-MapNet\shortcite{jiang2024pmapnet} & C+SD & 44.8 & 30.6 & 45.6 & 40.3 \\
SuperFusion\shortcite{dong2024superfusion} & C+L & 35.6 & 22.8 & 39.4 & 32.6 \\
CRFusion\shortcite{guan2026crfusion} &
C+L & 36.8 & 23.8 & 41.7 & 34.1 \\
Driver2Map(ours) & C+Sat & 48.3 & 29.8 & 46.3 & 41.6 \\
Driver2Map(ours) & C+SD+Sat & \textbf{49.7} & \textbf{32.6} & \textbf{48.0} & \textbf{43.4} \\
\hline
\end{tabular}
\caption{Results on the nuScenes validation set under two BEV ranges in IoU metrics. \textbf{C}, \textbf{SD}, \textbf{Sat}, and \textbf{L} denote onboard multi-view camera images, standard-definition maps, satellite images, and LiDAR clouds, respectively.}
\label{tab:iou_merged_6col}
\end{table}

\subsection{Results}
\label{sec:results}

\subsubsection{Comparison with SOTA Methods in AP Metrics.} 

As shown in Table~\ref{tab:ap}, Driver2Map achieves the best performance with the full \textbf{C+SD+Sat} input, validating the effectiveness of our multi-modal fusion strategy as well as the importance of complementary information from the SD map and satellite image. When using only the \textbf{C+Sat} input, Driver2Map significantly outperforms SatforHDMap. This is because our proposed PPMR module leverages the pre-trained HD map structural priors to effectively fill in gaps in the initial predictions caused by dynamic occlusions. Furthermore, with \textbf{C+SD} as input, Driver2Map also exceeds P-MapNet and SDTagNet. This indicates that our PGBF effectively mitigates cross-view interference; even without satellite priors, our Driver2Map yields more stable BEV representations, thereby enhancing overall prediction performance.

\subsubsection{Comparison with SOTA Methods in IoU Metrics.} As shown in Table~\ref{tab:iou_merged_6col}, we compare the IoU performance of the map semantic segmentation task under two BEV ranges. Our method yields consistent improvements across all three categories. Under the $60\,\text{m} \times 30\,\text{m}$ setting, Driver2Map with only \textbf{C+Sat} still outperforms HDMapNet+SatforHDMap by $5.6$ mIoU, showing that our fusion strategy can better exploit complementary information between satellite images and onboard multi-view camera images.


Finally, on both IoU and AP metrics, Driver2Map achieves significant improvements in the \textit{Crossing} category, which can be attributed to our PGBF module. \textit{Boundary} and \textit{Divider} are extensible along the road, making it highly probable that each camera captures information about these elements. In contrast, a complete \textit{Crossing} element can only exist either in front of or behind a vehicle. Thus, our PGBF reduces interference from irrelevant information by leveraging camera orientation and distance, effectively mitigating prediction challenges associated with such objects.

\begin{table}[h]
\centering

\renewcommand{\arraystretch}{1.10}
\small

\begin{tabular}{
l
@{\hspace{5pt}}c@{\hspace{5pt}}  
c@{\hspace{5pt}}                
|@{\hspace{4pt}}c@{\hspace{4pt}} 
|@{\hspace{4pt}}c@{\hspace{4pt}}                 
}

\toprule
\textbf{Setting} & \textbf{mIoU} & \textbf{mAP} &
\textbf{IoU (D/C/B)} & \textbf{AP (D/C/B)} \\
\midrule
\multicolumn{5}{l}{\textit{Module}}\\
\quad w/o PGBF & 44.0  & 57.5  &
46.1/38.7/47.2 & 51.3/58.9/62.1 \\
\quad w/o PPMR & 53.1  & 62.1  &
55.1/53.0/51.4 & 56.2/63.1/67.0 \\
\midrule
\multicolumn{5}{l}{\textit{Inputs}}\\
\quad w/o SAT & 49.2  & 59.7  &
50.2/48.0/49.3 & 52.7/60.6/65.7 \\
\quad w/o SD & 50.0  & 60.2  &
51.5/48.0/50.4 & 53.8/61.5/65.4 \\
\midrule
\textbf{Driver2Map} & \textbf{54.0}  & \textbf{62.9} &
\textbf{56.7/53.4/51.9} & \textbf{57.1/64.0/67.5} \\
\bottomrule
\end{tabular}
\caption{Ablation study on modules and inputs. \textbf{PGBF}: Pose-Guided BEV Fusion. \textbf{PPMR}: Pretrained Prior for Map Refinement.
\textbf{SAT}: satellite image. \textbf{SD}: standard-definition map. \textbf{D/C/B} denotes Divider/Crossing/Boundary.}
\label{tab:ablation_1}
\end{table}

\subsection{Ablation Study}

We conduct extensive ablation studies to systematically evaluate the effect of each component, input data and hyperparameters in our model. All experiments are performed on the nuScenes validation set under a $60\,\text{m} \times 30\,\text{m}$ BEV range.

\begin{figure*}[t]
  \centering
  \includegraphics[height=9cm]{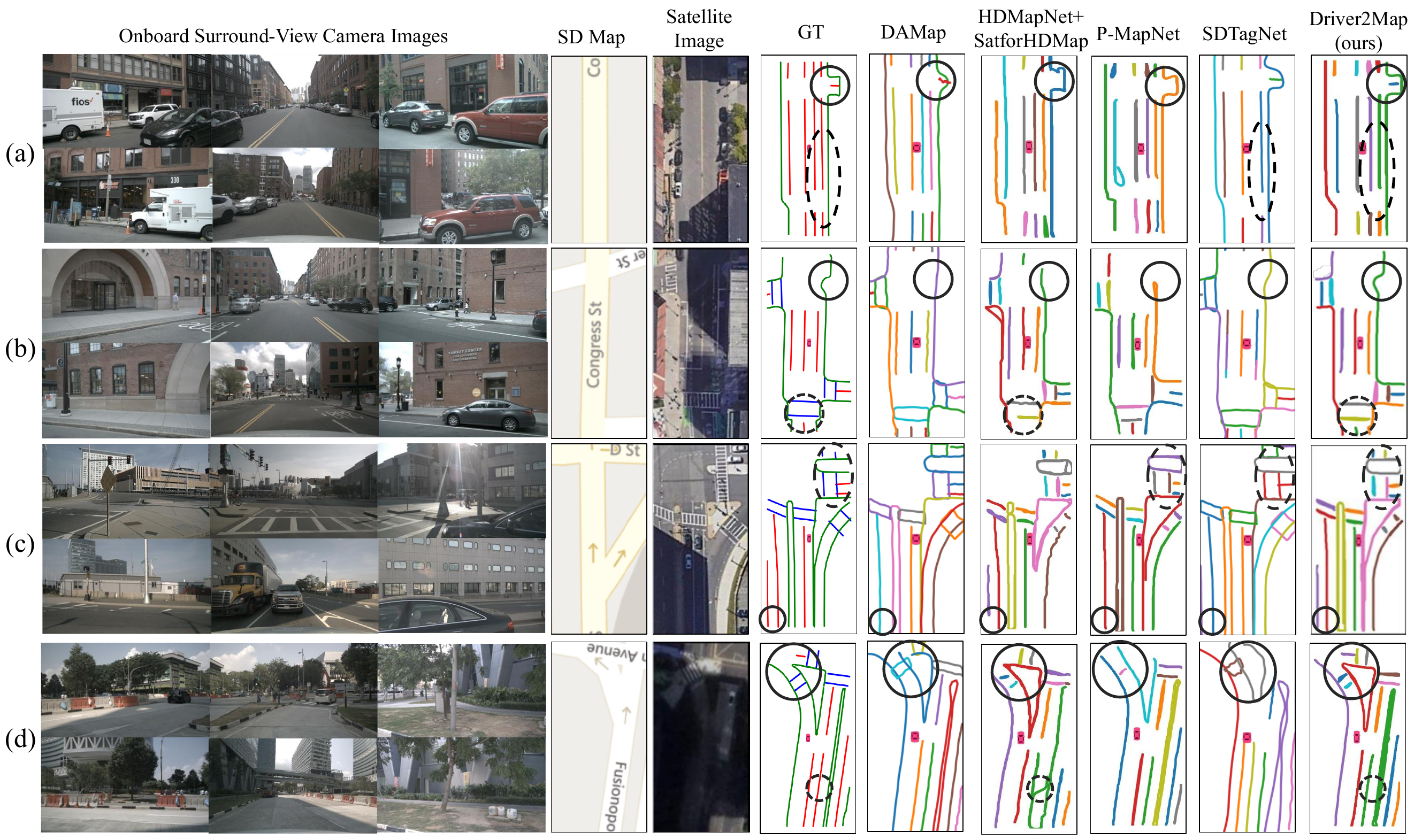}
  \caption{We compare our predictions with other methods within a $60\,\mathrm{m} \times 30\,\mathrm{m}$ BEV range. In (c), DAMap and P-MapNet fail to recover the lane line in the circled region. SDTagNet predicts this lane line but with a clear spatial shift, likely due to errors in the SD map. SatforHDMap detects the lane line with satellite image prior, but the result is incomplete because of shadow occlusion. In contrast, our method accurately recovers both its location and complete structure, demonstrating the complementarity of SD and satellite priors and the ability of the ``Pretrained Prior for Map Refinement'' module to recover occluded map elements.}
  \label{fig:vis_result}
\end{figure*}

\subsubsection{Ablations on modules and inputs.} Taking the complete Driver2Map as the baseline, we selectively remove individual modules or input priors and report the resulting IoU and AP performance in Table~\ref{tab:ablation_1}. Overall, the results confirm that both our proposed module and input priors deliver distinct and complementary advantages. Notably, removing PGBF causes the largest degradation, dropping mIoU by $10.0$ and mAP by $5.4$. The decline is particularly evident on the challenging \emph{Crossing} category (IoU: $53.4\rightarrow38.7$, AP: $64.0\rightarrow58.9$), validating that PGBF effectively mitigates multi-view feature overlap and improves instance consistency by letting each map element be dominated by the most relevant camera views. Additionally, both satellite image and SD map inputs provide important complementary priors: removing either one leads to declines across all metrics. Furthermore, we observe that removing the SAT input results in a more significant drop in mIoU compared to removing the SD input. Since IoU is a pixel-level metric that is more sensitive to local details, without the SAT input, the model lacks the process of detail enhancement, leading to a greater IoU decline.

\begin{table}[h]
\centering

\fontsize{9pt}{12pt}\selectfont\setlength{\tabcolsep}{1mm}{
\begin{tabular}{c|cc@{\hspace{0.5em}}c|cc}
\toprule
$\alpha/\beta$ & mIoU & IoU(D/C/B) &
$\alpha/\beta$ & mIoU & IoU(D/C/B) \\
\midrule
1:1 & 53.37 & 55.77/52.96/51.39 &
     &      &                 \\
\textbf{1:2} & \textbf{54.02} & \textbf{56.72/53.41/51.93} &
2:1 & 53.24 & 55.54/52.96/51.22 \\
1:3 & 53.46 & 55.93/52.95/51.51 &
3:1 & 53.11 & 55.33/52.94/51.05 \\
1:4 & 53.47 & 55.95/52.95/51.52 &
4:1 & 52.98 & 55.13/52.90/50.90 \\
1:5 & 53.48 & 55.96/52.95/51.53 &
5:1 & 52.86 & 54.95/52.86/50.76 \\
\bottomrule
\end{tabular}
}

\caption{Effect of the $\alpha:\beta$ ratio on IoU performance. In this table,  \textbf{D/C/B} denotes Divider/Crossing/Boundary. Moreover, we observe that the IoU performance saturates when $\alpha:\beta>5:1$ or $\alpha:\beta<1:5$. Therefore, these results are omitted from the table due to space limitations.
}
\label{tab:alpha_beta}
\end{table}

\subsubsection{Ablations on hyperparameters.} 

 We investigate the effect of the ratio between the balancing coefficients $\alpha$ and $\beta$ in the ``Pose-Guided BEV Fusion'' module proposed in Section~\ref{sec:image-to-bev} (see Eq.~\ref{eq:w_cam}), where the two coefficients respectively control the distance and orientation weights. As shown in Table~\ref{tab:alpha_beta}, the model achieves the best performance when $\alpha:\beta = 1:2$, and remains relatively stable under nearby settings, indicating that our method is not particularly sensitive to these hyperparameters. In particular, settings with $\alpha<\beta$ generally outperform those with $\alpha>\beta$, suggesting that orientation is more informative than distance in the proposed module.

\subsection{Qualitative Visualization}

We visualize the prediction results on the nuScenes validation set under a BEV range of $60\,\mathrm{m} \times 30\,\mathrm{m}$, and compare our method with DAMap \cite{dong2025damap} (camera-only), SatforHDMap \cite{gao2024satforhdmap} (camera with satellite prior), and P-MapNet \cite{jiang2024pmapnet} and SDTagNet \cite{immel2026sdtagnet} (camera with SD prior). 
We selected complex intersection scenarios for evaluation. As illustrated in \Cref{fig:vis_result}, our approach yields accurate predictions for both the overall road topology and fine-grained details. This is because the input SD map provides essential road skeleton and structural cues, enabling our model to correctly predict under complex scenarios such as intersections, while the satellite imagery supplements missing details, allowing precise delineation of lane markings, curves and other fine elements.



\section{Conclusion}


In this paper, we propose \textbf{Driver2Map}, a human-inspired online HD map construction framework that integrates onboard multi-view images, SD maps, and satellite images. It employs a two-stage alignment strategy to reduce cross-modal spatial misalignment, a ``Pose-Guided BEV Fusion'' module to mitigate multi-view feature overlap, and a ``Pretrained Prior for Map Refinement'' module to recover incomplete map elements under dynamic occlusions. In future work, we will explore the application of the constructed paired SD maps and satellite imagery to other autonomous driving tasks.



\bibliography{aaai2027}



\clearpage
\appendix

\section*{Appendix}

We provide more detailed descriptions of the proposed method, implementation details, and additional quantitative and qualitative results in the Appendix.
\begin{figure*}[t]
    \centering
    \includegraphics[width=\textwidth]{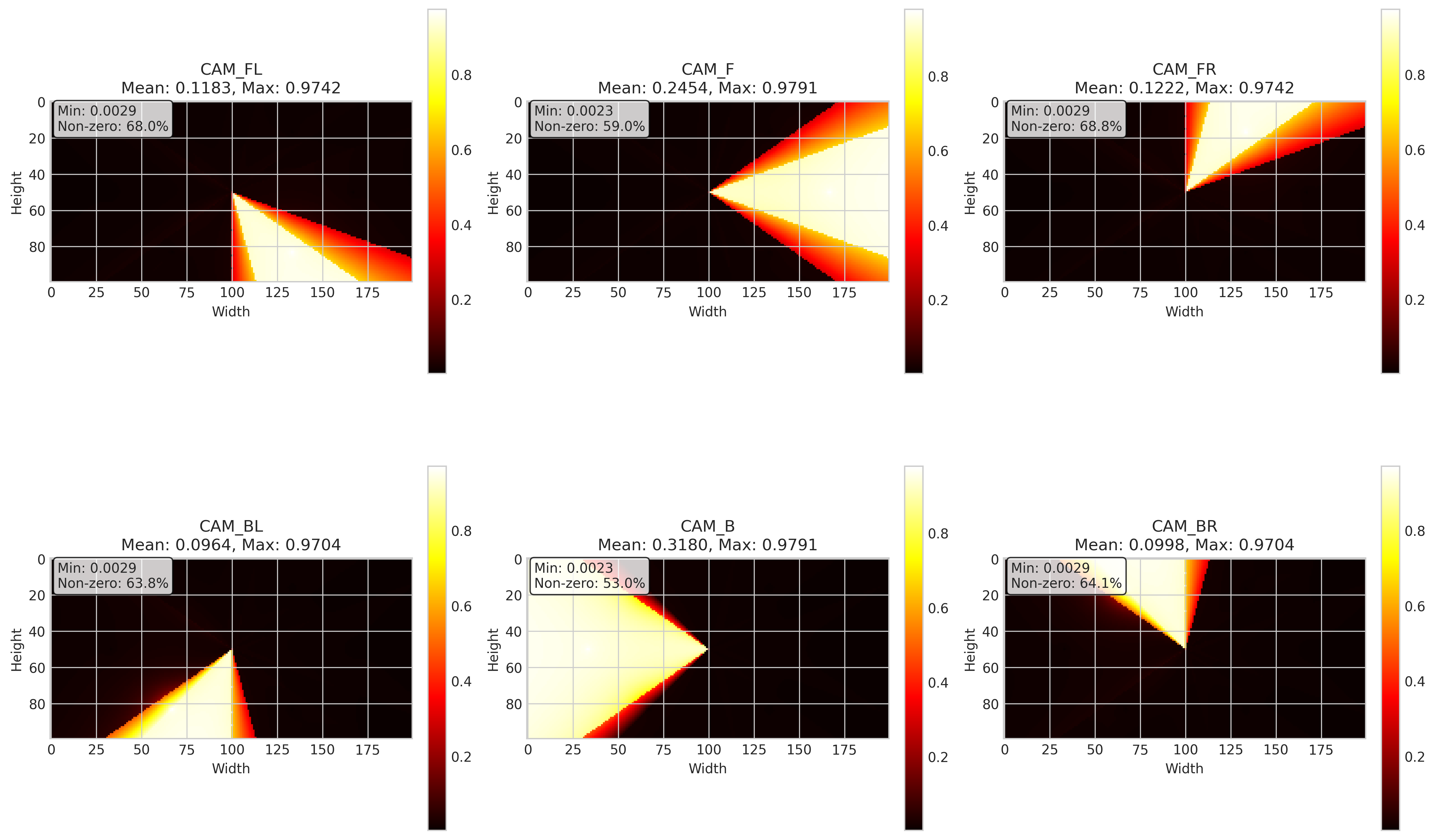}
    \caption{Visualization of the six camera-weight maps generated by
    PGBF for a representative sample from the nuScenes dataset. Brighter
    colors indicate larger normalized contributions to BEV feature
    aggregation, whereas dark regions are intentionally suppressed
    because they lie outside, or far from, the dominant viewing region
    of the corresponding camera. The reported mean is calculated over
    the complete BEV raster and is therefore not a measure of camera
    confidence or feature quality. The visualization follows a
    $60\,\mathrm{m}\times30\,\mathrm{m}$ BEV range, with the ego vehicle
    located at the center and its front pointing to the right; hence,
    the right and left halves of the figure correspond to the front and
    rear regions, respectively. 
    Owing to the nuScenes camera-coordinate
    convention and its conversion to the displayed BEV raster
    coordinates, the visual positions of \texttt{CAM\_FL} and
    \texttt{CAM\_FR}, as well as those of \texttt{CAM\_BL} and
    \texttt{CAM\_BR}, appear reversed. This is an expected consequence
    of the coordinate transformation \textbf{rather than an incorrect camera
    assignment.}
    }
    \label{fig:camera_weight_visualization}
\end{figure*}

\section{More Explanations about Our Method}

\subsection{Visualization of pose-guided camera weights}
Figure~\ref{fig:camera_weight_visualization} visualizes the spatial
camera-weight maps produced by the Pose-Guided BEV Fusion (PGBF)
module. As described in the main paper, the weight assigned to each
camera is jointly determined by its orientation, field of view, and
the distance of a BEV location from the corresponding camera sector.
The contributions of all six cameras are then normalized at each BEV
location before feature aggregation. Consequently, these weights
represent the \emph{relative contribution} of each view at a given
location, rather than prediction confidence or feature quality. A low
weight outside a camera's dominant viewing region is therefore an
intended behavior: it suppresses geometrically irrelevant features and
reduces cross-view interference, while smooth responses near adjacent
fields of view preserve complementary information from neighboring
cameras. Moreover, the mean values shown in the figure are computed
over the entire BEV raster, including locations outside each camera's
effective field of view, and hence should not be interpreted as weak
camera utilization. Since the weighting pattern is constructed from
camera geometry instead of scene appearance or dataset-specific
semantic cues, PGBF can be directly adapted to different camera
configurations by updating the corresponding calibration parameters.

\subsection{Detailed Procedure of Coarse Alignment}
\label{appendix:coarse_alignment}

\begin{figure*}[t]
  \centering
  \includegraphics[height=7cm]{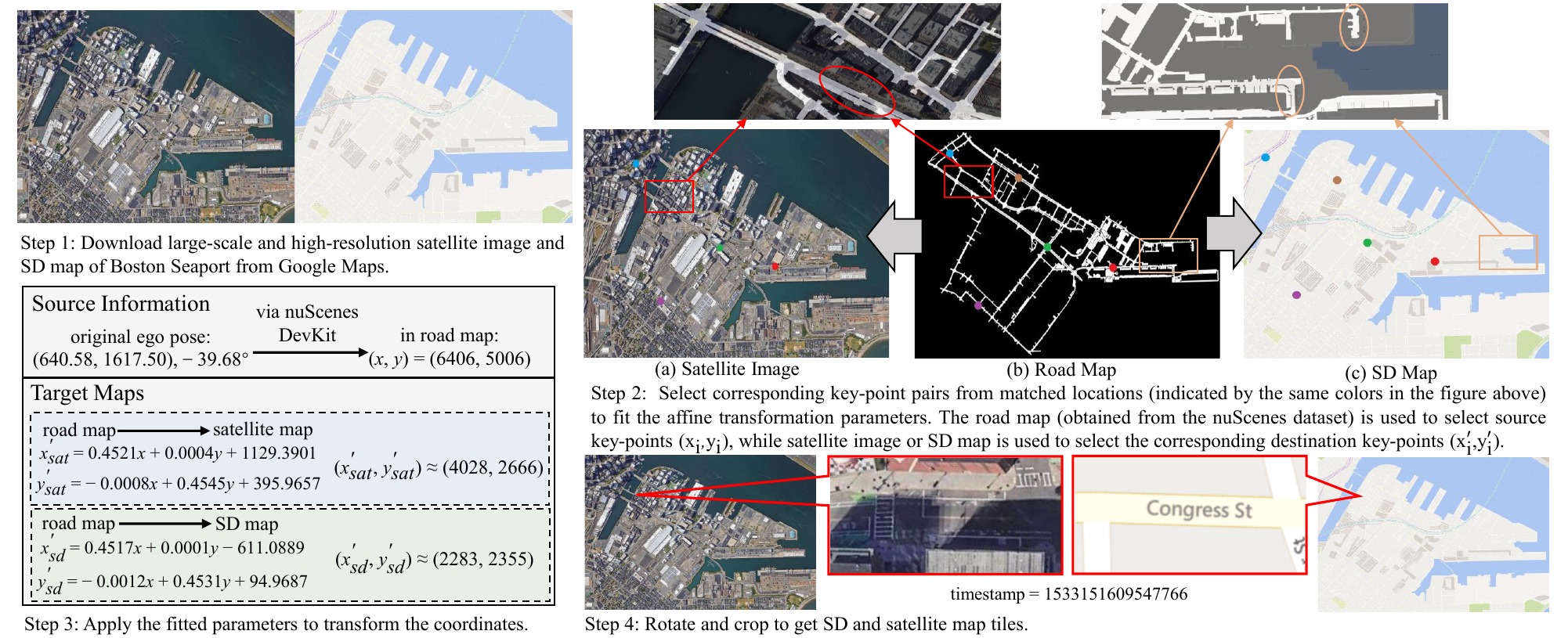}
  \caption{Detailed workflow for generating coarsely aligned SD map and
  satellite image priors, illustrated using Boston Seaport.}
  \label{fig:vis_coarse}
\end{figure*}

Figure~\ref{fig:vis_coarse} provides a detailed
illustration of the coarse-alignment procedure introduced in
Section~3.3 and Algorithm~1 of the main paper. The objective of coarse
alignment is to establish a city-level geometric correspondence among
the nuScenes road map, the SD map, and the satellite image before model
training. These maps are obtained from different sources (nuScenes, Google Maps, Google Earth) and are
represented in different coordinate systems and resolutions. Therefore,
the same physical location may have different pixel coordinates in the
three maps. Directly cropping the prior maps using the original ego pose
would consequently produce spatially mismatched SD and satellite tiles.

In Step~1, we obtain city-scale, high-resolution satellite imagery and
the corresponding SD map for each nuScenes region. These maps cover the
complete region of interest and serve as the source maps from which
sample-specific local priors are subsequently generated. The nuScenes
road map is adopted as the reference coordinate system because each
driving sample is already associated with an ego pose
$p_i=(x_i,y_i,\theta_i)$ in the nuScenes coordinate system.

In Step~2, we establish landmark correspondences between the nuScenes
road map and each external prior map. Specifically, we manually identify
the same physical landmarks, such as road intersections, road endpoints,
bridge structures, and distinctive road boundaries, in the road map and
in the satellite image or SD map. The landmarks indicated by the same
color in Figure~\ref{fig:vis_coarse} - Step 2 form a corresponding
key-point pair. For each region, we use $N=50$ key-point pairs to
estimate the transformation. The road-map coordinates are treated as
source points, while their corresponding coordinates in the satellite
image or SD map are treated as destination points. Since the satellite
image and SD map have different coordinate systems, two affine mappings
are fitted independently rather than assuming that the two priors are
already mutually aligned.

For a source point $\mathbf{q}=[x,y,1]^\top$ in the road-map coordinate
system, its corresponding point in prior map $m\in\{\mathrm{sat},
\mathrm{sd}\}$ is computed as
\begin{equation}
\begin{bmatrix}
x'_m\\
y'_m
\end{bmatrix}
=
\mathbf{A}_m
\begin{bmatrix}
x\\
y\\
1
\end{bmatrix},
\quad
\mathbf{A}_m=
\left[
\begin{array}{ccc}
a_m & b_m & e_m\\
c_m & d_m & f_m
\end{array}
\right].
\label{eq:appendix_affine_alignment}
\end{equation}
The parameters of $\mathbf{A}_{\mathrm{sat}}$ and
$\mathbf{A}_{\mathrm{sd}}$ are estimated separately by least squares
over their respective key-point pairs. A full affine transformation is
used because it can jointly account for the dominant translation,
rotation, and scale discrepancies between maps. The fitted mappings are
region-specific but sample-independent: they are estimated once during
data preparation and then reused for all samples from the corresponding
region.

In Step~3, the ego pose of each sample is first converted by the
nuScenes DevKit into the pixel coordinate of the nuScenes road map.
Equation~\ref{eq:appendix_affine_alignment} is then applied to this
road-map coordinate using $\mathbf{A}_{\mathrm{sat}}$ and
$\mathbf{A}_{\mathrm{sd}}$, yielding the corresponding crop centers in
the city-scale satellite image and SD map, respectively. The numerical
example in Figure~\ref{fig:vis_coarse} shows this
conversion for one timestamp: the same road-map position is mapped to
different pixel positions in the two target maps because their origins,
resolutions, and coordinate conventions differ.

Finally, in Step~4, we use the transformed centers together with the ego
heading $\theta_i$ to rotate and crop the city-scale prior maps,
producing a pair of local $100\times200$ satellite and SD-map tiles.
The rotation places the two tiles in the ego-centric orientation used by
the target BEV representation. Consequently, the generated tiles
describe the same local road region as the onboard observations and can
be paired with the corresponding nuScenes sample during training and
inference.

This coarse alignment removes the dominant city-level discrepancies
between coordinate systems and makes large-scale prior-map generation
stable and reusable. Nevertheless, manual landmark selection, map
rasterization, and local map distortions may leave small spatial
residuals. These residual errors are handled separately by the
learnable fine-alignment module inside the Detail Enhancement stage.
Thus, coarse alignment provides global geometric correspondence during
data preparation, while fine alignment performs local feature-level
correction within the network.

\section{More Quantitative Results}

\subsection{Inference Efficiency}
\label{appendix:inference_efficiency}

We further evaluate the inference efficiency of Driver2Map and several
representative online HD map construction methods. As reported in
Table~\ref{tab:inference_efficiency}, Driver2Map achieves
$11.10$ FPS on a single RTX 4090 GPU with a batch size of 1,
corresponding to approximately $90.1$ ms per frame. Its throughput is
comparable to P-MapNet and slightly lower than SatforHDMap, despite
jointly incorporating onboard multi-view images, an SD map, and a
satellite image. This result indicates that the additional topology
enhancement, detail enhancement, and map refinement operations introduce
only moderate computational overhead. Although Driver2Map does not
achieve the highest FPS among the compared methods, its throughput is
sufficient for online HD map construction while providing the accuracy
improvements reported in the main paper.
\begin{table}[t]
    \centering
    \caption{Inference-efficiency comparison on a single NVIDIA RTX
    4090 GPU. All methods are evaluated with a batch size of 1. We first
    perform 50 warm-up iterations and then report the average FPS over
    250 inference iterations. Higher FPS indicates better efficiency.}
    \label{tab:inference_efficiency}
    \setlength{\tabcolsep}{10pt}
    \begin{tabular}{lc}
        \toprule
        \textbf{Method} & \textbf{Average FPS} \\
        \midrule
        SatforHDMap~\shortcite{gao2024satforhdmap}       & 11.52 \\
        P-MapNet~\shortcite{jiang2024pmapnet}          & 11.10 \\
        SDTagNet~\shortcite{immel2026sdtagnet}          &  8.13 \\
        DAMap~\shortcite{dong2025damap}             & \textbf{14.18} \\
        \textbf{Driver2Map (ours)} & 11.10 \\
        \bottomrule
    \end{tabular}
\end{table}

\begin{figure*}[h]
    \centering
    \includegraphics[height=11cm]{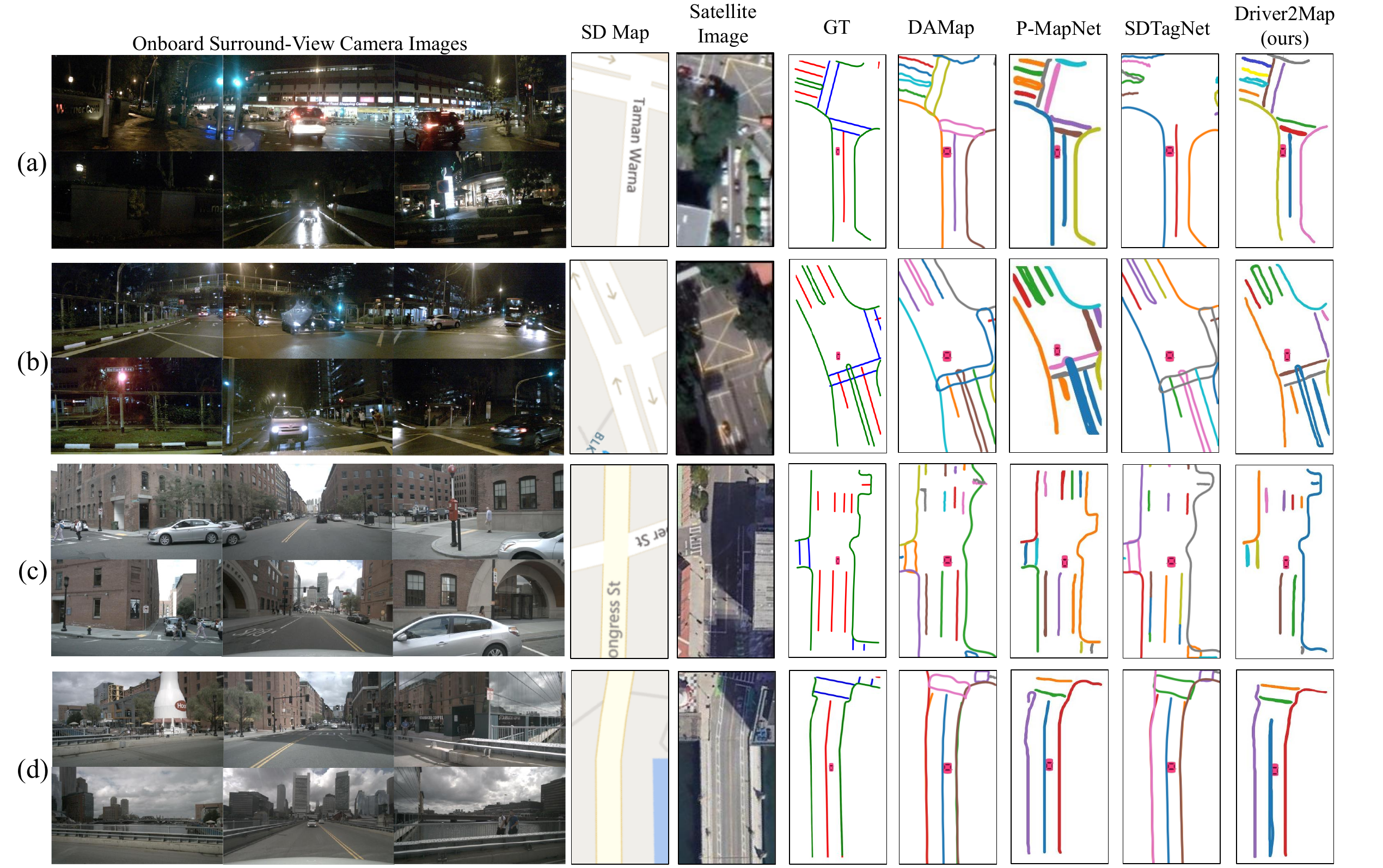}
    \caption{
    Additional qualitative results on the nuScenes validation set.
    Columns show onboard surround-view camera images, the SD map, the
    satellite image, ground-truth HD maps, and predictions from DAMap,
    P-MapNet, SDTagNet, and Driver2Map. Rows (a) and (b) correspond to
    nighttime scenes under poor illumination, where onboard images are
    affected by darkness, glare, and reflections. Despite these
    degraded observations, Driver2Map preserves the road topology and
    recovers fine map elements by combining the structural prior from
    the SD map with the complementary geometric information from the
    satellite image. Rows (c) and (d) show representative failure
    cases. The dominant road layout is largely recovered, but some
    thin lane dividers, short lane segments, or sharply curved
    boundaries are incomplete or slightly misaligned because of
    residual cross-modal registration errors, satellite-image
    occlusion, and ambiguity in the camera observations.
}
\label{fig:additional_visualization}
\end{figure*}

\section{More Qualitative Visualization}

\paragraph{Performance under poor illumination.}
Figures~\ref{fig:additional_visualization}(a) and
\ref{fig:additional_visualization}(b) show two nighttime scenes with
severely degraded onboard camera observations. Low illumination,
headlight glare, reflections, and dark image regions make lane markings
and road boundaries difficult to identify from camera images alone. In
these cases, the SD map provides a stable topological description of
the road layout, while the satellite image supplies complementary
bird's-eye-view geometry that is less sensitive to nighttime
appearance. By progressively incorporating these two priors, Driver2Map
maintains the overall road topology and recovers most lane-level
structures despite the poor visual quality of the onboard images. The
results are consistent with the motivation in the introduction and with
the qualitative comparisons in the main paper: external priors
compensate for the limitations of vehicle-view perception under adverse
illumination, while the pretrained map prior further helps complete
structures that are partially missing from the current observation.

\paragraph{Failure cases and limitations.}
Figures~\ref{fig:additional_visualization}(c) and
\ref{fig:additional_visualization}(d) present representative failure
cases. Driver2Map generally preserves the dominant road topology in
these examples, but some fine-grained structures are incomplete,
slightly displaced, or merged with neighboring map elements. Such
errors are particularly visible for thin and closely spaced lane
dividers, short lane segments near intersections, and road boundaries
with abrupt curvature changes. These structures may be weakly expressed
in the SD map, partially occluded or distorted by shadows and buildings
in the satellite image, and ambiguous in the onboard views because of
perspective and dynamic objects. In addition, residual local
misalignment after coarse map registration can affect the fusion of
fine details, while the PPMR module is designed to restore plausible
missing structures rather than to resolve every instance-level
ambiguity. Therefore, the failure cases indicate the remaining
limitations of cross-modal alignment and fine-detail localization,
rather than a failure to recover the underlying large-scale road
topology.

\section{Discussions}

\paragraph{Limitations and future work.}
Our current coarse-alignment procedure requires selecting landmark
corresponding points for each city during data preparation, which limits its
scalability to unseen regions. In practical deployment, this step could
be replaced by an automated geo-referencing pipeline that obtains the
vehicle location and heading in real time and queries SD maps and
satellite imagery through map software APIs. More broadly, existing online HD
map construction methods, including ours, are primarily evaluated in
dataset-based inference settings and still remain at a laboratory
prototype stage rather than a fully deployed engineering system. Future
work will therefore investigate automatic cross-source registration,
robust API-based prior retrieval and caching, and efficient adaptation
to changing map coverage, coordinate conventions, and real-world
latency constraints.


\end{document}

%% file: CoarseAlignment.tex
\begin{algorithm}[h]
\caption{Coarse Alignment for Prior Generation}
\label{alg:coarse_alignment}
\begin{algorithmic}[1]
\Require city road map $\mathcal{M}_{road}$, ego pose $p_i=(x_i,y_i,\theta_i)$
\Ensure prior tiles $\mathcal{T}_{sd}$ and $\mathcal{T}_{sat}$ for $p_i$

\Statex \textbf{// Step 1: Acquire city-scale prior maps}
\State $\mathcal{I}_{sd}, \mathcal{I}_{sat} \leftarrow \mathrm{DownloadMaps} (\mathrm{region})$

\Statex \textbf{// Step 2: Select keypoint pairs (shown in Figure \ref{fig:keypoint})}
\State $\mathcal{K}_{sd} \leftarrow \{(q_u, q_u^{sd})\}_{u=1}^{N}$, \quad
       $q_u \in \mathcal{M}_{road},\; q_u^{sd} \in \mathcal{M}_{sd}$
\State $\mathcal{K}_{sat} \leftarrow \{(q_u, q_u^{sat})\}_{u=1}^{N}$, \quad
       $q_u \in \mathcal{M}_{road},\; q_u^{sat} \in \mathcal{I}_{sat}$

\Statex \textbf{// Step 3: Estimate road-to-prior affine mappings}
\State $f_{sd} \leftarrow \mathrm{FitAffine}(\mathcal{K}_{sd})$, $f_{sat} \leftarrow \mathrm{FitAffine}(\mathcal{K}_{sat})$

\Statex \textbf{// Step 4: Generate local prior tiles}
\State $\mathcal{T}_{sd} \leftarrow \mathrm{Crop}(\mathrm{Rotate}(\mathcal{I}_{sd},\theta_i), f_{sd}(p_i))$
\State $\mathcal{T}_{sat} \leftarrow \mathrm{Crop}(\mathrm{Rotate}(\mathcal{I}_{sat},\theta_i), f_{sat}(p_i))$
\State \Return $\mathcal{T}_{sd}, \mathcal{T}_{sat}$
\end{algorithmic}
\end{algorithm}